\documentclass[runningheads, letter]{llncs}

\usepackage{amssymb}
\usepackage{graphicx}
\usepackage{boldcommands}
\usepackage{url}
\usepackage{graphicx,graphics}
\usepackage{url}
\usepackage{amsmath,amsfonts,amssymb,textcomp,booktabs,threeparttable}
\usepackage{amssymb}
\usepackage{hhline}
\usepackage{multirow}
\usepackage{stmaryrd}
\usepackage{boldcommands}
\usepackage{tabularx}
\usepackage{booktabs}
\usepackage{braket}
\usepackage{algorithm}
\usepackage{algpseudocode}
\usepackage{tikz}
\usepackage[T1]{fontenc}
\usepackage{graphicx}
\usepackage{caption}

\urldef{\mailsa}\path|ajog@mgh.harvard.edu|
\newcommand{\keywords}[1]{\par\addvspace\baselineskip
\noindent\keywordname\enspace\ignorespaces#1}
\newcommand{\0}{\hspace*{1.1ex}}

\begin{document}

\mainmatter  

\title{Pulse Sequence Resilient Fast Brain Segmentation}

\titlerunning{Sequence Resilient Fast Brain Segmentation}

%
%
\author{Amod Jog\inst{1} \and Bruce Fischl\inst{1}}

\authorrunning{A.~Jog et al.}

\institute{Athinoula A. Martinos Center for Biomedical Imaging,
Massachusetts General Hospital and Harvard Medical School, Boston, MA, USA\\
\mailsa}
%
%

\toctitle{}
\tocauthor{}
\maketitle

\begin{abstract}
%
%
%
Accurate automatic segmentation of brain anatomy from $T_1$-weighted~($T_1$-w)
magnetic resonance images~(MRI) has been a computationally intensive bottleneck
in neuroimaging pipelines, with state-of-the-art results obtained by unsupervised intensity modeling-based methods and multi-atlas registration and label fusion. With the advent of powerful supervised convolutional neural networks~(CNN)-based learning algorithms, it is now possible to produce a high quality brain segmentation within seconds. However, the very supervised nature of these methods makes it difficult to generalize them on data different
from what they have been trained on. Modern neuroimaging studies are necessarily multi-center initiatives with a wide variety of acquisition protocols. Despite stringent protocol harmonization practices, it is not possible to standardize the whole gamut of MRI imaging parameters across
scanners, field strengths, receive coils etc., that affect image contrast. In this paper we propose a CNN-based segmentation
algorithm that, in addition to being highly accurate and fast, is also resilient to variation in the input $T_1$-w acquisition.
Our approach relies on building approximate forward models of $T_1$-w pulse sequences that produce a typical test image. We use the forward models to
augment the training data with test data specific training examples. These augmented data can be used to update and/or build a more robust segmentation model that is more attuned to the test data imaging  properties.  Our method generates highly accurate, state-of-the-art segmentation results~(overall Dice overlap=0.94), within seconds and is consistent across a wide-range of protocols.
\end{abstract}

\section{Introduction}
\label{sec:intro}
Whole brain segmentation is one of the most important tasks in a neuroimage processing pipeline. A segmentation output consists of labels for
white matter, cortex, and subcortical structures such as the thalamus, hippocampus, amygdala, and others. Structure volumes and shape statistics that rely on volumetric segmentation are regularly used to quantify differences between healthy and diseased populations~\cite{fischl2004aseg}.
An ideal segmentation algorithm of course needs to be highly accurate, but also, critically, it needs to be robust to variations in input data. Large modern MRI datasets are necessarily multi-center initiatives to gain access to a larger pool of subjects. It is very difficult to achieve perfect acquisition harmonization across different sites due to variations in scanner
manufacturers, field strengths, receive coils, pulse sequences, available contrast, and resolution. Variations in site-specific parameters introduce bias and increase variation in downstream image processing including segmentation~\cite{han2006ni,jovicich2013ni}.
Low computational load is a yet another desirable property of a segmentation algorithm.
A fast, memory-light segmentation algorithm enables quicker processing of large datasets and wider adoption.

Existing segmentation algorithms can be broadly classified into three types:
(a)~unsupervised, (b)~multi-atlas registration-based, and (c)~supervised.
Unsupervised algorithms~\cite{fischl2004aseg,shiee2010ni}
fit the observed intensities of the input image to an underlying atlas-based
model and perform maximum a posteriori labeling. They
assume a functional form~(e.g. Gaussian) of intensity distributions and results can degrade if the input distribution differs from this assumption. Efforts have been made to develop a hybrid approach~\cite{puonti2016samseg} that is robust to input sequences and also leverages manually labeled training data. Unsupervised methods are usually computationally intensive, taking 0.5--4 hours to run. Multi-atlas registration and label fusion~(MALF) algorithms achieve state-of-the-art~\cite{asman2012tmi,wang2013malf} segmentation accuracy. However, they require multiple computationally expensive registrations followed by label fusion. Registration quality can also suffer if the test image contrast properties are significantly different from the atlas images. Recently, with the success of deep learning methods in medical imaging, supervised segmentation approaches built on 3D CNNs have produced accurate segmentations with a runtime of a few seconds to minutes~\cite{niftynet17,wachinger2017}. Despite the powerful local and global context that these models provide, they are vulnerable to subtle contrast differences that are inevitably present in multi-scanner MRI studies. However, with appropriate training using
test-data specific augmentation, as we will show in Section~\ref{sec:results},
these differences can be essentially removed.

We present PSACNN--Pulse Sequence Augmented Convolutional Neural Network; a CNN-based multi-label segmentation approach that employs an augmentation scheme of generating training image patches on-the-fly that appear as if they have been imaged using the pulse sequence of the test data.
PSACNN training consists of three major steps: (1) Estimating test data pulse sequence parameters, (2) applying test data pulse sequence forward models to training data nuclear magnetic resonance~(NMR) parameters to create test data specific training features, (3) training a deep CNN to predict the segmentation using the augmentation. We will describe each of these steps in detail in Section~\ref{sec:method}.
\section{Method}
\label{sec:method}
\begin{figure}[!ht] \tabcolsep 1pt
	\begin{center}

	\begin{tabular}{cc@{}@{}}
		\raisebox{-\height}{\includegraphics[width=0.7\textwidth]{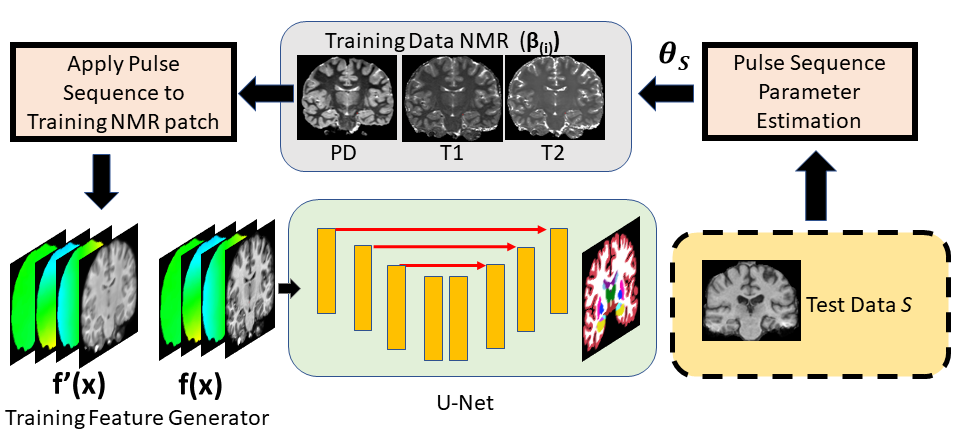}} &
		\begin{tabular}[t]{c}
			\raisebox{-\height}{\includegraphics[height=0.12\textwidth]{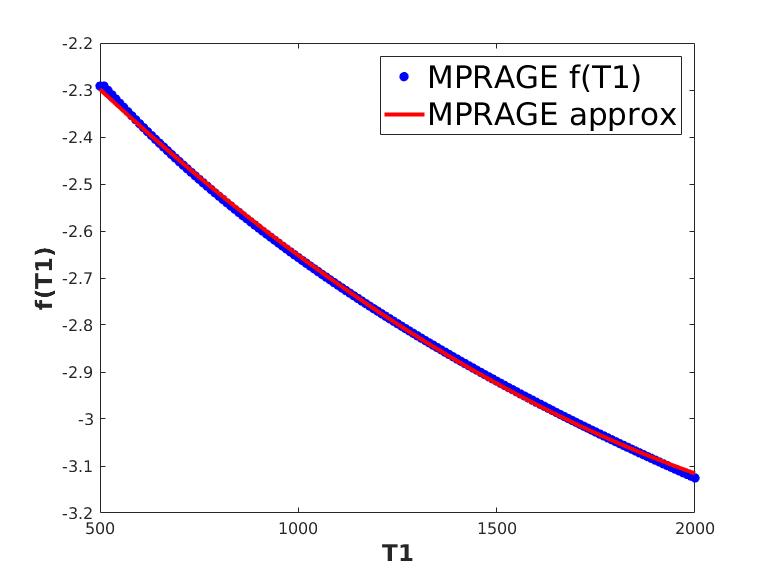}} \\
			(b)\\
			\raisebox{-\height}{\includegraphics[height=0.12\textwidth]{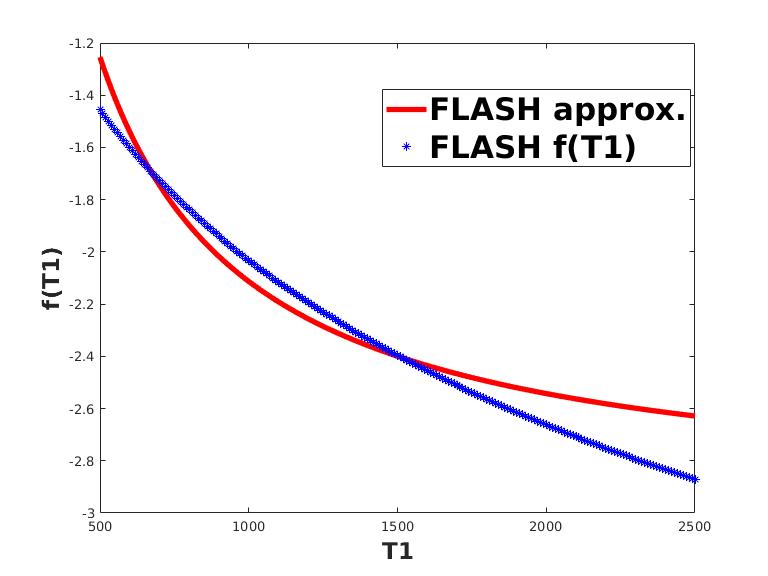}}\\
			(c)
		\end{tabular} \\
	    (a) & \qquad

	\end{tabular}
\end{center}
	\caption{(a) Workflow of test-data specific augmentation for training, (b) fit of $T_1$ component in the MPRAGE equation~(blue) and our approximation~(red), (c) fit of $T_1$ component~(blue) with our approximation~(red) for FLASH.}
	\label{fig:method}
\end{figure}
Let $\mathcal{A} = \{A_{(1)}, A_{(2)}, \ldots, A_{(M)}\}$ be a collection of $M$ $T_1$-w images with a corresponding expert manually labeled image set $\mathcal{Y} = \{Y_{(1)}, Y_{(2)}, \ldots Y_{(M)}\}$. The paired collection $\{\mathcal{A}, \mathcal{Y}\}$ is referred to as the atlas image set or the training image set. We assume that $\mathcal{A}$ is acquired using the
pulse sequence $\Gamma_{A}$, where $\Gamma_A$ can be MPRAGE~(magnetization prepared gradient echo)~\cite{mugler1990mrm} or FLASH~(fast low angle shot), SPGR~(spoiled gradient echo), and others.
In addition, let $\mathbf{\mathcal{B}}=\{\boldbeta_{(1)}, \boldbeta_{(2)}, \ldots, \boldbeta_{(M)}\}$ be the corresponding NMR parameter maps. For each $i \in \{1, \ldots, M\}$ we have $\boldbeta_{(i)} = [\rho_{(i)}, {T_1}_{(i)}, {T_2}_{(i)}]$, where $\rho_{(i)}$ is a map of proton densities, and ${T_1}_{(i)}$ and ${T_2}_{(i)}$ store the longitudinal~($T_1$), and transverse~($T_2$) relaxation time maps respectively. Most atlas datasets do not acquire or generate $\boldbeta$ maps. We generate them using image synthesis
for our atlas dataset using a previously acquired dataset~\cite{Fischl2004} that estimated $\rho$ and $T_1$ maps from multi-echo FLASH images. We describe how to complement existing atlas sets with these maps in the Supplementary Material.
\paragraph{\textbf{Step 1: Estimating Test Data Acquisition Parameters:}}
\phantomsection
\label{subsec:estimation}
Let $S$ be the test image we want to obtain the segmentation for, acquired via pulse sequence $\Gamma_S$. We assume that we have access to the test image $S$~(or test dataset) prior to training so that we can design test data-specific augmentations. We would like to estimate the pulse sequence parameters of $\Gamma_S$ directly from the image $S$. The intensity $S(\boldx)$ at voxel $\boldx$ is given by the imaging equation of $\Gamma_S$. Equation~\eqref{eq:flash}
shows the imaging equation for the FLASH sequence. $S(\boldx)$ is a function of the acquisition parameters repeat time~($TR$), echo time~($TE$), flip angle~($\alpha$), gain~($G$), and tissue NMR parameters $\boldbeta(\boldx) = [\rho(\boldx), T_1(\boldx), T_2^*(\boldx)]$.
\begin{equation}
S_{\textrm{FLASH}}(\boldx) = \Gamma_{\textrm{FLASH}}(\boldbeta(\boldx), [TR, TE, \alpha]) = G{\rho} \sin \alpha \frac{(1 - e^{-\frac{TR}{T_1}})}{1 - \cos\alpha e^{-\frac{TR}{T_1}}}e^{-\frac{TE}{T_{2}^{*}}}
\label{eq:flash}
\end{equation}
The MPRAGE sequence is more complex to model~\cite{wang2014mprage}. In general, it is difficult to derive an imaging equation for most pulse sequences, and given $\boldbeta(\boldx)$ and the acquisition parameter set, it is necessary to run a full Bloch equation simulation to obtain voxel intensities.  Additionally, the number of scanner parameters that can affect signal intensity is very large on modern MRI scanners and for many datasets detailed parameter sets may not be available. Therefore, we would like to estimate these directly
from the image intensities. However, it is generally not possible to robustly estimate all pulse sequence parameters purely from the image intensities themselves. Even when the equation is well-understood, the highly nonlinear dependence of intensities on imaging parameters makes their estimation unstable. Therefore, we have formulated approximations of the MPRAGE and FLASH equations with a smaller number of parameters that can be estimated directly and robustly from the image intensities. Our approximation for the FLASH sequence is shown in Eqns.~\eqref{eq:flash_approx1}-\eqref{eq:flash_approx2}.
 \begin{align}
 \log(S_{\textrm{FLASH}}) & = \log(G\sin\alpha) + \log({\rho}) + \log(\frac{(1 - e^{-\frac{TR}{T_1}})}{1 - \cos\alpha e^{-\frac{TR}{T_1}}}) -\frac{TE}{T_{2}^{*}} \label{eq:flash_approx1}  \\
 &\approx \theta_0 + \log(\rho) + \frac{\theta_1}{T_1} + \frac{\theta_2}{T_2},
 \label{eq:flash_approx2}
 \end{align}
 where $\boldtheta_{\textrm{FLASH}} = \{\theta_0, \theta_1, \theta_2\}$ forms our parameter set. For the range of values of $T_1$ in the human brain $(500, 3000)$~ms at 1.5~T, our approximation fits the signal dependence on $T_1$ well~(see Figs.~\ref{fig:method}(b) and~\ref{fig:method}(c)). Equation~\eqref{eq:mprage_approx} is our approximation for the MPRAGE imaging equation provided by Wang et al.~\cite{wang2014mprage}.
 	\begin{equation}
 	\log(S_{\textrm{MPRAGE}}) \approx \theta_0 + \log(\rho) + \theta_1T_1 + \theta_2{T_1}^2.
 	\label{eq:mprage_approx}
 	\end{equation}
Given test image $S$ we estimate $\boldtheta_S$ by a strategy that is similar to Jog et al.~\cite{jog2015media}. Let $S_c$, $S_g$, $S_w$ be the mean intensities of cerebrospinal fluid~(CSF), gray matter~(GM), and white matter~(WM), respectively. These can be obtained by a three-class classification scheme by fitting a Gaussian mixture model to the intensity distribution. Let $\boldbeta_{c}$, $\boldbeta_{g}$, $\boldbeta_{w}$ be the mean NMR values for CSF, GM, and WM classes, respectively, that are obtained from previously acquired data~\cite{Fischl2004}. Thus, we have a system of three linear equations $S_c = \Gamma_S(\boldbeta_{c}; \boldtheta_S); S_g = \Gamma_S(\boldbeta_g; \boldtheta_S); S_w = \Gamma_S(\boldbeta_w; \boldtheta_S)$ and three unknown parameters in $\boldtheta_S$ that can be solved to obtain the estimate $\hat{\boldtheta}_S$~(see Fig.~\ref{fig:method}(a)).
\paragraph{\textbf{Step 2: Feature Extraction and Augmentation:}}
We use the U-Net CNN architecture~\cite{unet2015} as our predictor.
GPU memory constraints prevent us from using the $256\times256\times256$ brain image as an input. For each $A_{(i)}$ in $\mathcal{A}$, we sample a $32\times32\times32$-sized patch $\boldp(\boldx)$ at voxel $\boldx$. We observed that training a purely patch-based U-Net leads to segmentation errors due to incorrect localization. To provide more global context, we use information from the nonlinear warp produced by the FreeSurfer pipeline~\cite{Fischl2004} that aligns the FreeSurfer atlas to the test image $S$. For each voxel $\boldx$ in $S$,
we extract the FreeSurfer atlas 3D coordinates $\boldw(\boldx)$,  that warp to $\boldx$,
and produce $\boldw_p(\boldx)$--a $32\times32\times32\times3$-sized patch of warped FreeSurfer atlas coordinates. The final feature vector $\boldf(\boldx) = [\boldp(\boldx), \boldw_p(\boldx)]$, is a concatenation
of the intensity patch and the coordinates~(see Fig.~\ref{fig:method}(a)
2D patches shown for illustration).
Test data-specific augmented patches are generated by extracting NMR parameter patches from the $i^{\textrm{th}}$ NMR map for each $i \in \{1, \ldots, M\}$, $\boldbeta_{(i)} \in \mathcal{B}$ where the NMR patch at voxel $\boldx$ is denoted by $[\boldrho_{(i)}(\boldx), {\boldT_{1}}_{(i)}(\boldx),{\boldT_{2}}_{(i)}(\boldx)]$. The augmented patches are given by:
\begin{equation}
\boldp'(\boldx) = \Gamma_S([\boldrho_{(i)}(\boldx), {\boldT_{1}}_{(i)}(\boldx),{\boldT_{2}}_{(i)}(\boldx)]; \hat{\boldtheta}_{S})
\end{equation}
In addition, we also sample the 3D parameter space of $\boldtheta_S$ and generate non-test-data specific augmented patches to increase the breadth
of available training. The augmented patches are also concatenated with the warped FreeSurfer atlas coordinate patches $\boldw_p(\boldx)$ to generate the augmented features $\boldf'(\boldx) = [\boldp'(\boldx), \boldw_p(\boldx)]$ and added to the training~(Fig~\ref{fig:method}(a)).

Original and augmented features with corresponding manually labeled
image patches are used to train the U-Net, which has three pooling and corresponding upsampling stages~(see Figure~\ref{fig:method}(a)). Each orange block consists of two 3D convolutional layers followed by a batch normalization layer. The first block has 32 $3\times3\times3$-sized filters and the subsequent encoding blocks have twice the number of filters as the previous block. The red arrows signify skip connections where data from the encoding layers is concatenated as input to the decoding layers. All activations are ReLu except for the last layer that has a softmax activation. The output is $32\times32\times32\times L$ where $L=44$ is the number of labels. We use the Adam optimization algorithm to minimize soft Dice-based loss calculated
over all labels. The batch size is 32 and is divided equally between augmented
and original training features.
The total size of unaugmented training was $\approx$ $10^6$ patches from 16 subjects in $\mathcal{A}$. Validation data was generated from three subjects. During prediction, we extract features from test image $S$ and apply the trained U-Net to generate overlapping patches of label probabilities. These are combined together by averaging in the overlapping regions to produce the final probability map. The label with the highest probability is selected as the label in the hard segmentation.
\section{Experiments and Results}
\label{sec:results}
\begin{figure}[!h] \tabcolsep 1pt

\begin{tabular}{cc@{}@{}}
		\begin{tabular}[t]{@{}cc@{}}
		\raisebox{-\height}{\includegraphics[width=0.2\textwidth]{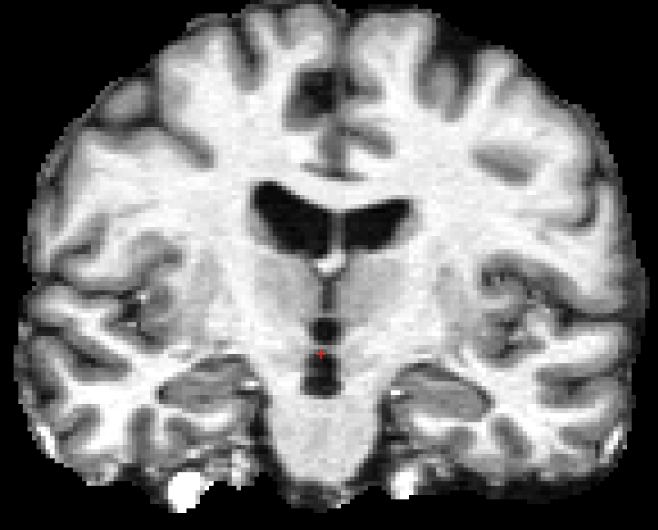}} &
		\raisebox{-\height}{\includegraphics[width=0.2\textwidth]{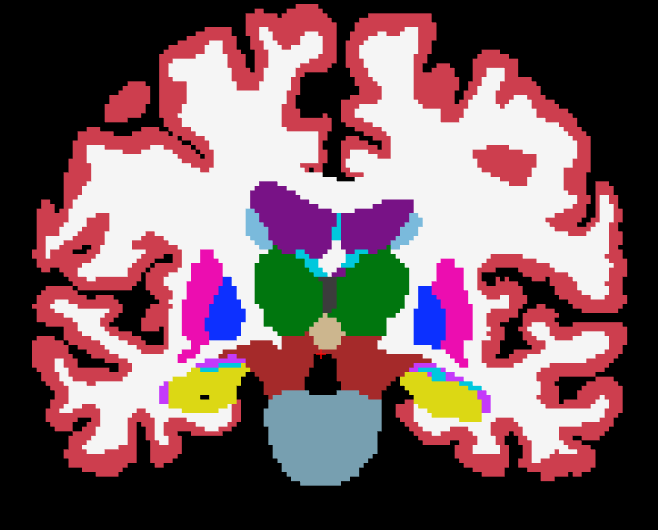}} \\
		MPRAGE & Manual \\[0cm]
		\qquad &
		\raisebox{-\height}{\includegraphics[width=0.2\textwidth]{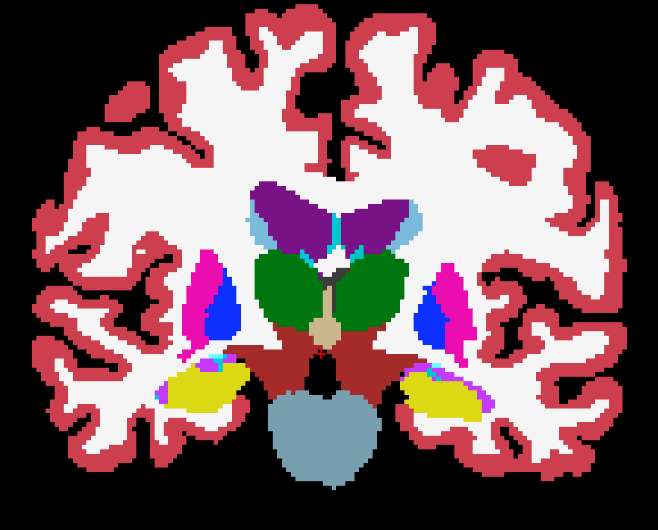}}\\
		\qquad & PSACNN
	\end{tabular} &
	\raisebox{-\height}{\includegraphics[width=0.6\textwidth]{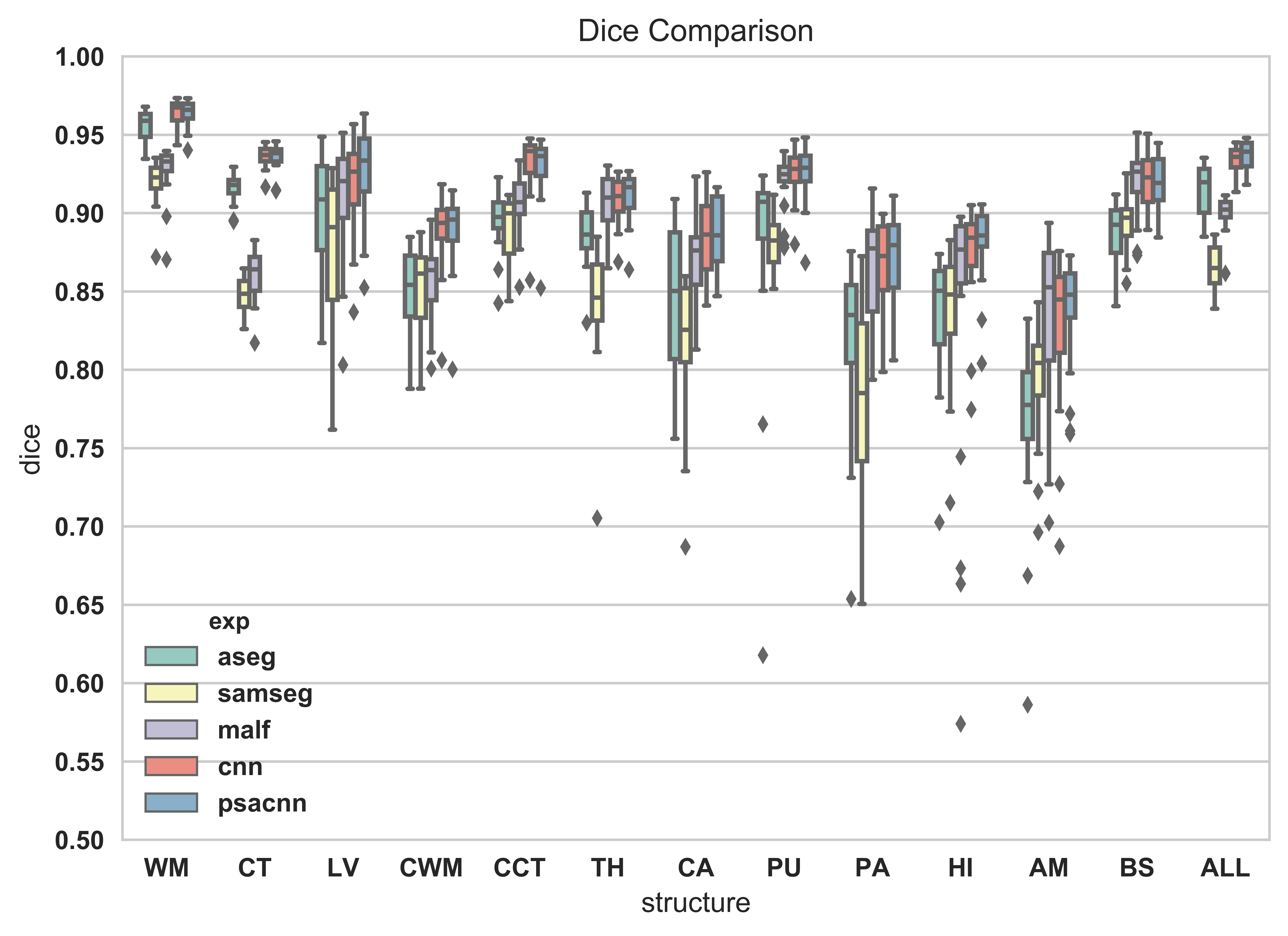}}
\end{tabular}
	\caption{Acronyms: white matter~(WM), cortex~(CT), lateral ventricle~(LV), cerebellar white matter~(CWM), cerebellar cortex~(CCT), thalamus~(TH), caudate~(CA), putamen~(PU), pallidum~(PA), hippocampus~(HI), amygdala~(AM), brain-stem~(BS), overlap for all structures~(ALL).}
\label{fig:bucknerseg}
\end{figure}

\paragraph{\textbf{3.1: Same Scanner Dataset}}
\label{sec:buckner}
In this experiment we compare the performance of PSACNN against unaugmented CNN~(CNN), FreeSurfer segmentation~(ASEG)~\cite{fischl2004aseg}, SAMSEG~\cite{puonti2016samseg}, and MALF~\cite{wang2013malf}, on test data with the same acquisition protocol as the training data. The complete dataset consists of 39 subjects with 1.5~T Siemens Vision MPRAGE acquisitions~(TR=9.7~ms, TE=4~ms TI=20~ms, voxel-size=$1\times1\times1.5$~mm$^3$) with expert manual labels done as per the protocol described in~\cite{buckner2004protocol}. We chose a subset of 16 subjects as training data for CNN, PSACNN, and as the atlas set for MALF. We used 3 subjects for validation for the CNNs, and 20 for testing. Figure~\ref{fig:bucknerseg} shows the results for all the algorithms and an example segmentation produced by PSACNN. CNN~(red) and PSACNN~(blue) have significantly higher Dice overlap~(ALL Dice $=0.9376$) than the next best method as tested using a paired t-test~($p < 0.01$) for most structures.

\paragraph{\textbf{3.2: Different Scanner Datasets}}
\begin{figure}[!t] \tabcolsep 1pt
	\centerline{
		\begin{tabular}{ccccc}
			\includegraphics[width=.2\textwidth]{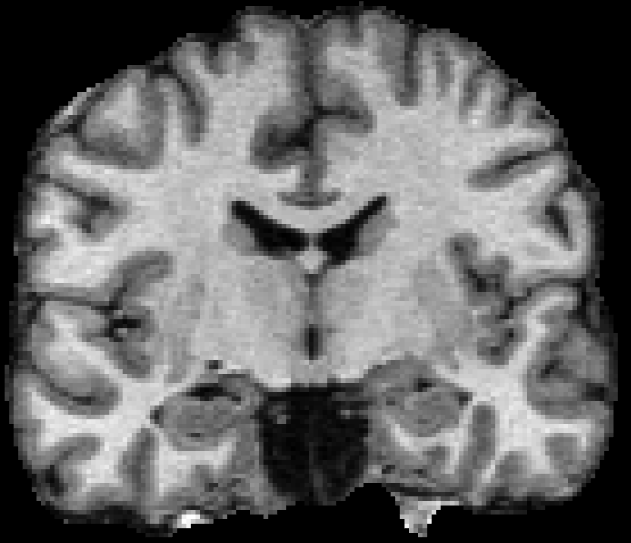} &
			\includegraphics[width=.2\textwidth]{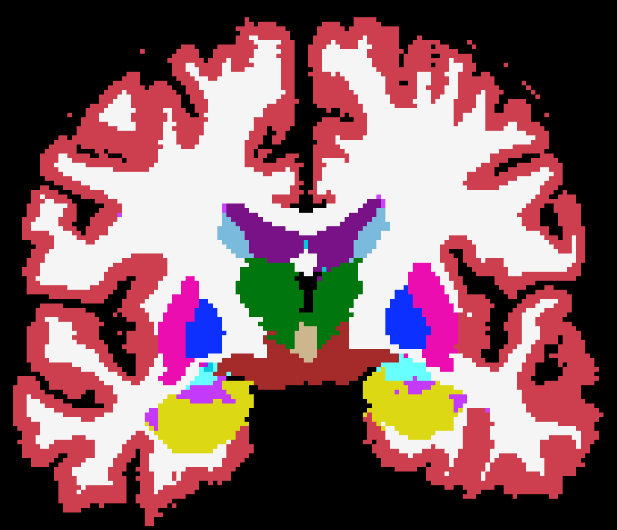} &
			\includegraphics[width=.2\textwidth]{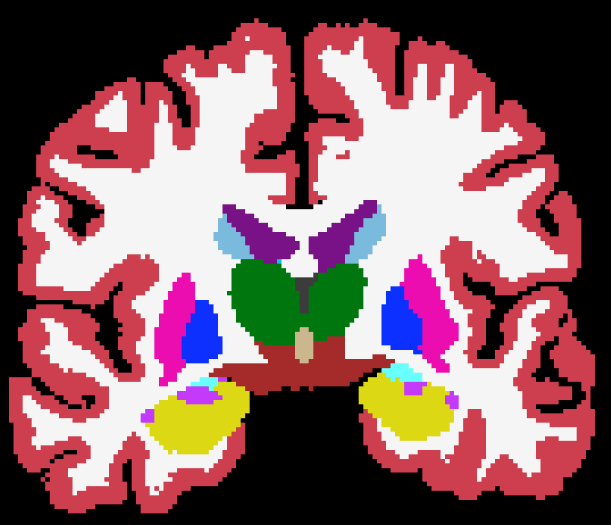} &
			\includegraphics[width=.2\textwidth]{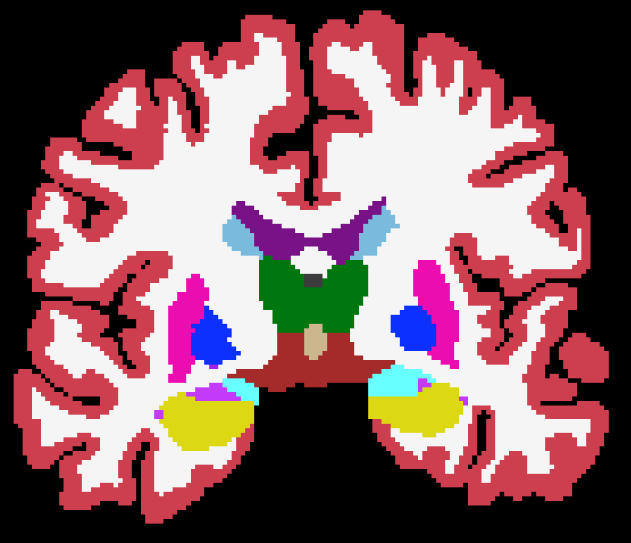}&
			\includegraphics[width=.2\textwidth]{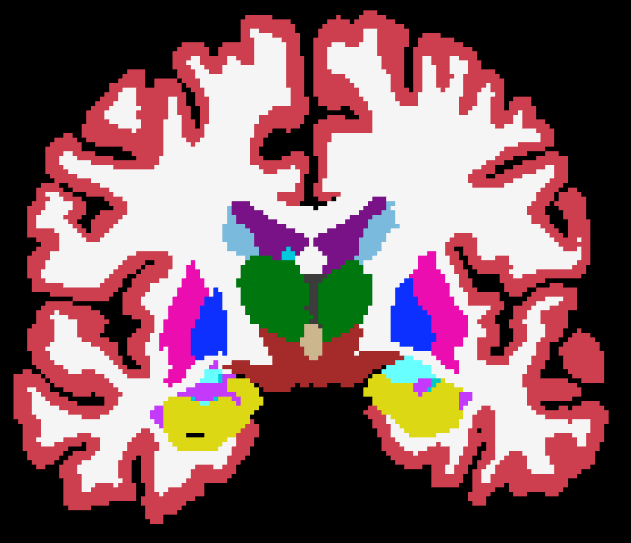}
			\\
			Siemens MPRAGE & SAMSEG & MALF & PSACNN & Manual\\
			\includegraphics[width=.2\textwidth]{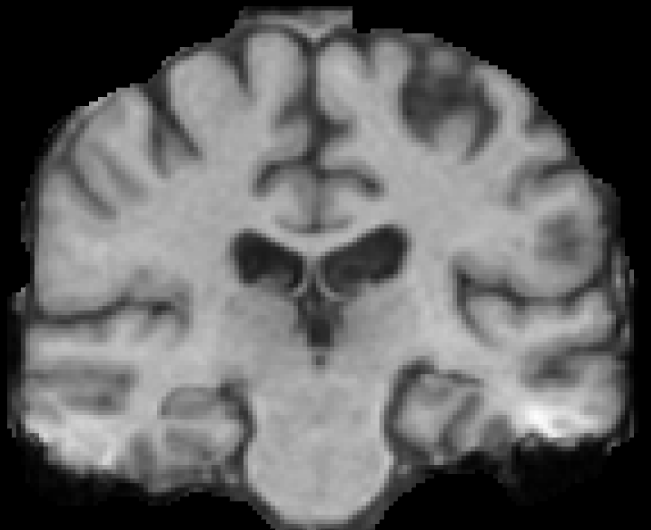} &
			\includegraphics[width=.2\textwidth]{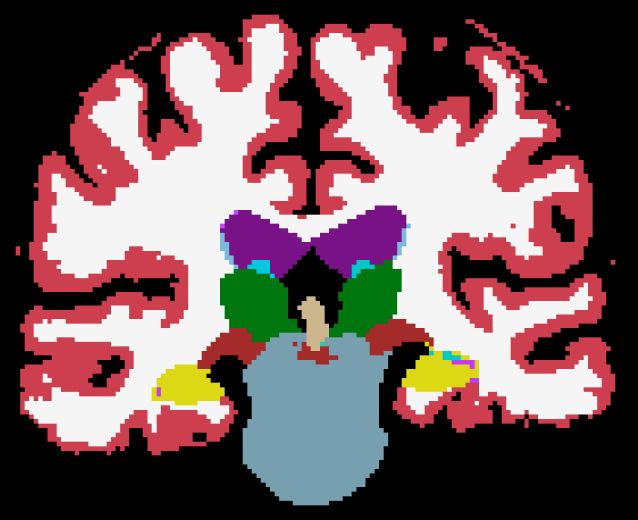} &
			\includegraphics[width=.2\textwidth]{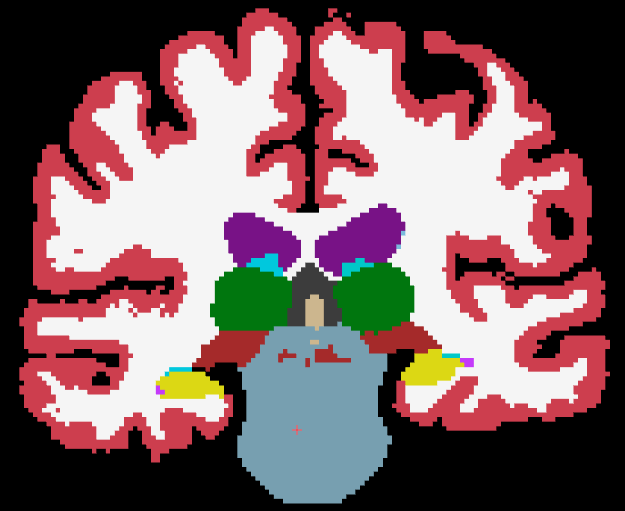} &
			\includegraphics[width=.2\textwidth]{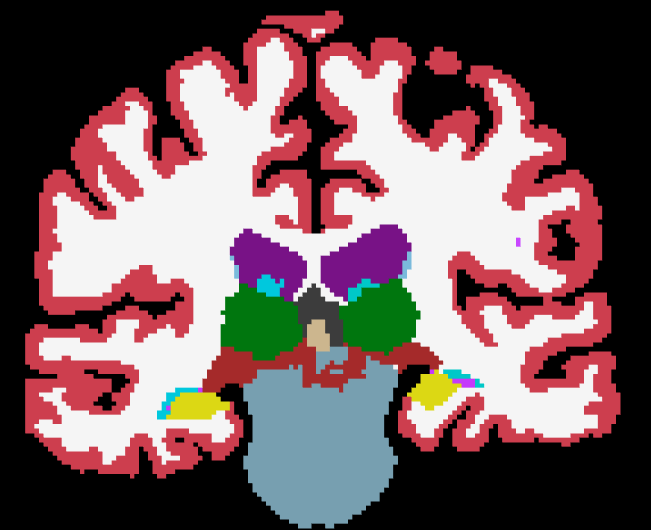}
			&
			\includegraphics[width=.2\textwidth]{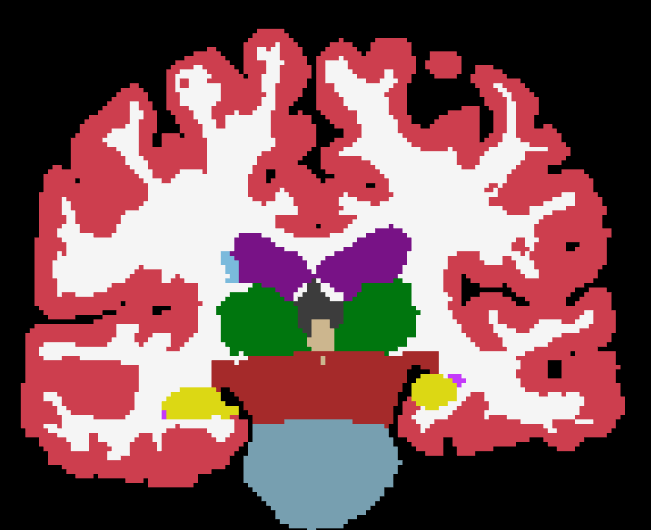}\\
			GE SPGR & SAMSEG & MALF & PSACNN & Manual\\
	\end{tabular}
	}
	\caption{Input Siemens and GE acquisitions with segmentation results from SAMSEG, MALF, and PSACNN, along with manual segmentations.}
	\label{fig:ge_siemens_seg}
\end{figure}

\begin{figure}[t] \tabcolsep 1pt
	\centerline{
		\begin{tabular}{cc}
			\includegraphics[width=.5\textwidth]{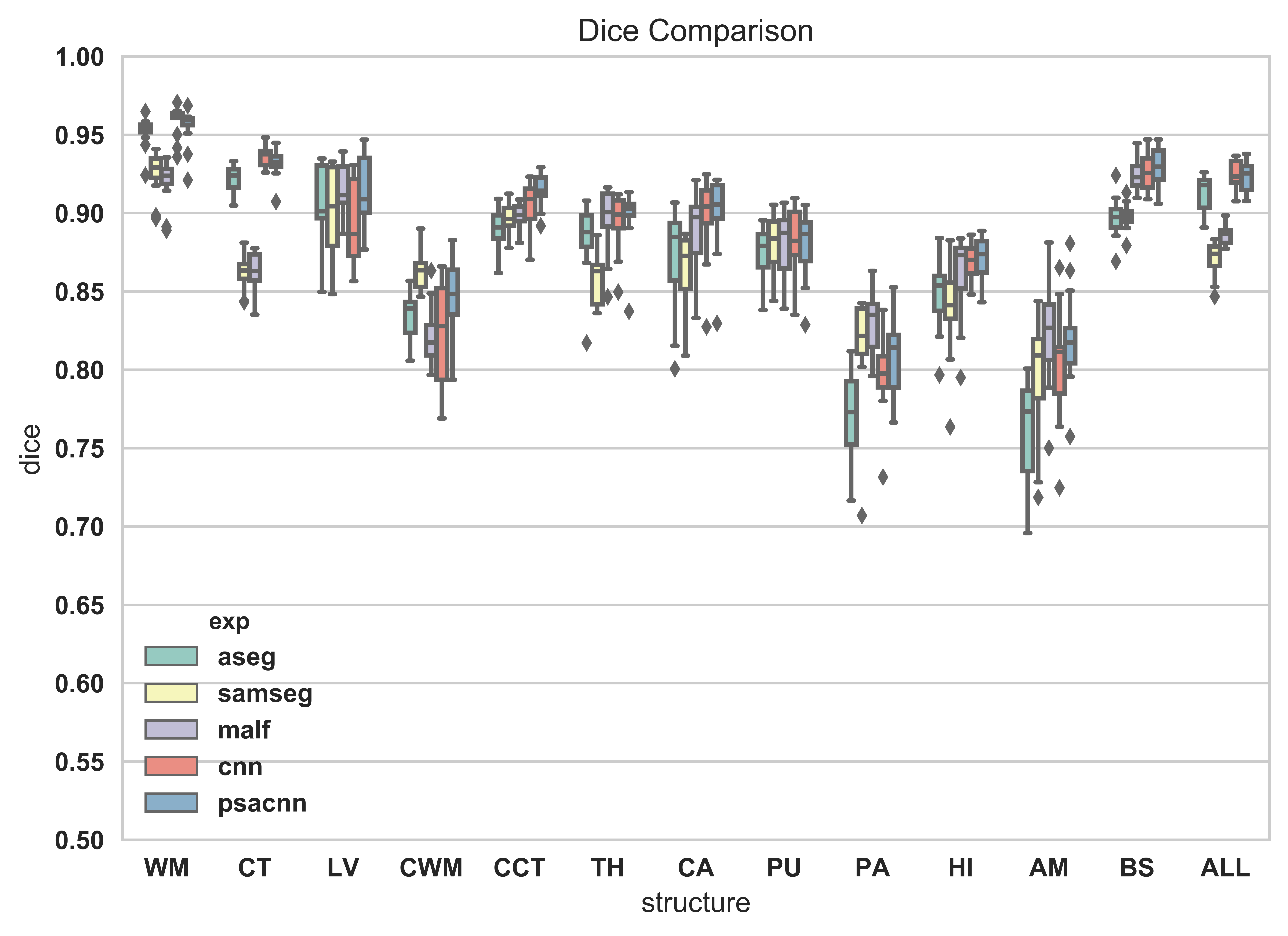} &
			\includegraphics[width=.5\textwidth]{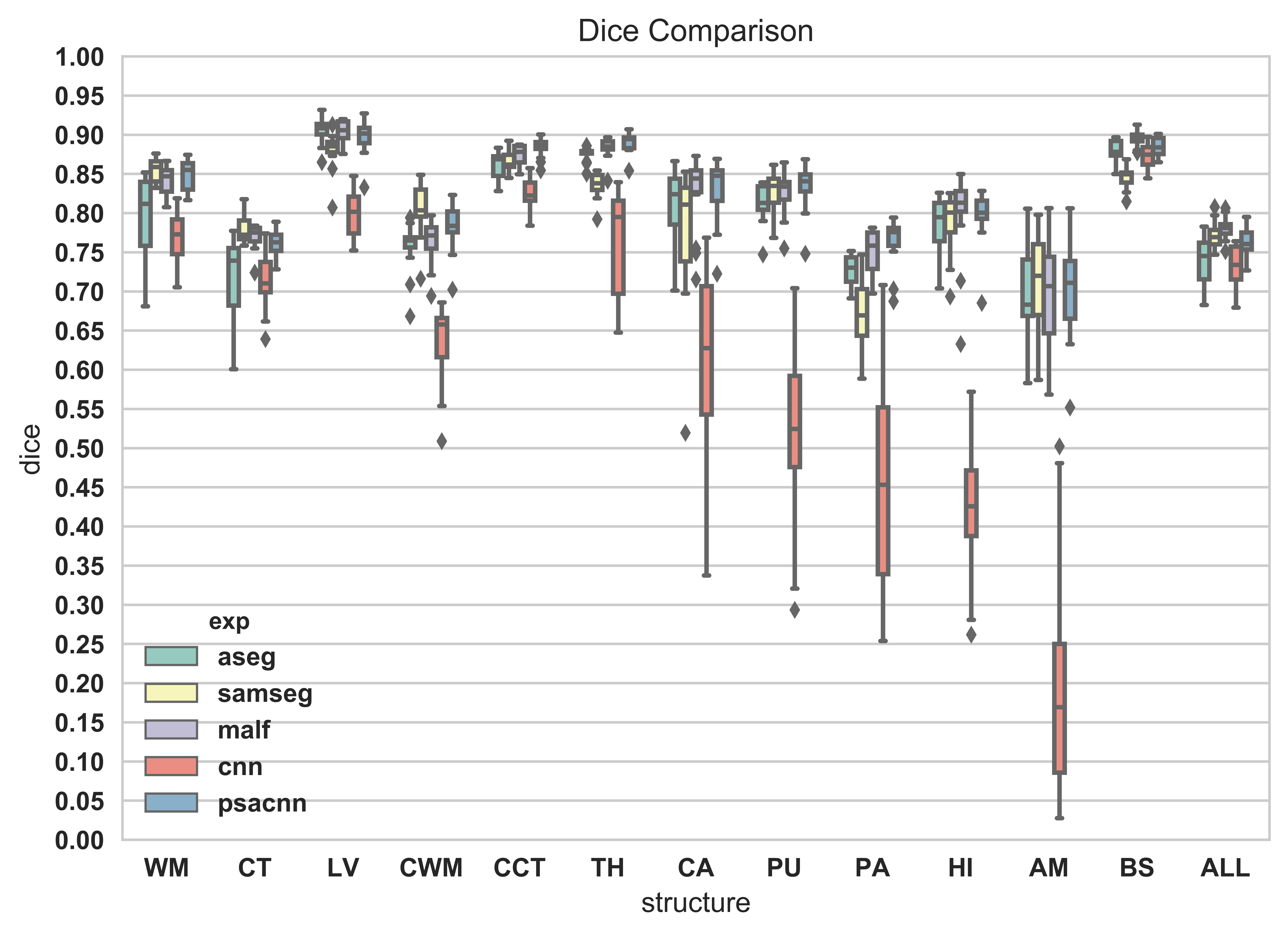}\\
			(a) & (b)
		\end{tabular}
	}
	\caption{Dice evaluations on (a) Siemens dataset, (b) GE dataset. For acronyms refer to Fig.~\ref{fig:bucknerseg} caption.}
	\label{fig:ge_siemens_dice}
\end{figure}

In this experiment we compare the accuracy of PSACNN against other methods on two datasets; (a) Siemens dataset: 14 subject MPRAGE scans acquired on a 1.5~T
Siemens SONATA scanner with the same pulse sequence as the training data, and (b) GE dataset: 13 subject SPGR~(spoiled gradient recalled) scans acquired on a 1.5~T GE Signa~(TR=35~ms, TE=5~ms, $\alpha=45^{\circ}$, voxel size=$0.9375\times0.9375\times1.5$~mm$^3$) scanner. Both datasets have expert manual segmentations generated with the same protocol as the training data. Figure~\ref{fig:ge_siemens_dice} shows comparison of Dice coefficients for all the methods. On the Siemens dataset CNN and PSACNN are significantly better than the rest~(Fig.~\ref{fig:ge_siemens_dice}(a)), as is expected due to its similarity with the training data. However, on the GE scans, which present a noticeably different tissue contrast~(Fig.~\ref{fig:ge_siemens_seg}), all methods show a reduced overlap~(Fig.~\ref{fig:ge_siemens_dice}(b)), but CNN has the worst performance of all as it is unable to generalize. PSACNN~(ALL Dice=$0.7636$) on the other hand, is robust to the contrast change due to pulse sequence-based augmentation, and produces segmentations that are comparable to the state-of-the-art algorithms such as SAMSEG~(ALL Dice=$0.7708$) and MALF~(ALL Dice=$0.7804$) in accuracy, with an order of magnitude lower processing time.
\paragraph{\textbf{3.3: Multi-scanner Consistency}}
In this experiment we tested the consistency of the segmentation produced by the various methods on four datasets acquired  from 15 subjects; \textit{MEF}: multi-echo FLASH acquired on Siemens 3~T TRIO scanner, \textit{TRIO}: MPRAGE acquired on Siemens TRIO, \textit{GE}: MPRAGE on 1.5~T GE Signa, and \textit{Siemens}: MPRAGE on 1.5~T Siemens SONATA scanner. \textit{Siemens} scan parameters are closest to that of the training dataset and we calculated absolute difference in structure volumes obtained
using different segmentation methods on datasets \textit{MEF}, \textit{TRIO}, \textit{GE} with respect to \textit{Siemens}.
In Table~\ref{tab:multiscanner_volumes} we report only WM volume differences due to lack of space and also because total white matter volumes are a good indicator of accumulated effects of pulse sequence variation.
 MALF shows the highest consistency for \textit{GE} and is second best for the rest. PSACNN provides the best performance for \textit{TRIO} and second best for the \textit{GE} dataset.
\begin{table}[!t]
	\caption{Mean and standard deviation~(Std. Dev.) of absolute WM volume differences between datasets \textit{MEF}, \textit{TRIO}, \textit{GE}, with \textit{Siemens} for different algorithms.}
	\label{tab:multiscanner_volumes} \tabcolsep 0pt
	\vspace*{0.6em}
	\centerline{
		\begin{tabular}{l c c c c c c}  \toprule  & \hspace*{6.0ex} &
			\multicolumn{1}{c}{\textit{MEF}} & \hspace*{4.0ex} &
			\multicolumn{1}{c}{\textit{TRIO}}& \hspace*{4.0ex} &
			\multicolumn{1}{c}{\textit{GE}}\\ && \textbf{Mean} \textbf{~(Std)} &&
			\textbf{Mean} \textbf{~(Std.)} && \textbf{Mean} \textbf{~(Std)}\\ \cmidrule{3-3}
			\cmidrule{5-5} \cmidrule{7-7}
			 ASEG && 21.38 (19.16) && \0{}2.48 (2.72) && \0{}9.13 (15.64) \\
			 SAMSEG && \0{}2.22 (\0{}1.26)$^*$ && \0{}2.12 (2.78) && \0{}7.49 (11.53) \\
			 MALF && \0{}4.14 (\0{}1.64) && \0{}2.18 (3.00) && \0{}4.33	(4.71)$^*$ \\
			 CNN && 87.13 (12.14) && \0{}2.18 (2.61) && \0{}9.02 (17.70) \\
			 PSACNN && \0{}7.13 (\0{}4.40) && \0{}1.65 (2.65)$^*$ && \0{}6.56 (10.68) \\
			\bottomrule
		 \end{tabular}
		  }
	\vspace{0.5em}
	\centerline{
		\scriptsize{ \begin{minipage}{0.93\textwidth} * Statistically significantly better than the next best method ($p  < 0.01$) using a paired t-test. \end{minipage} } }
\end{table}
\section{Discussion and Conclusions}
We have described PSACNN, a CNN-based segmentation algorithm that can adapt itself to the test data by generating test-data specific augmentation using an approximate forward model of MRI image formation. The augmentation can be used to robustly train or fine-tune any underlying predictor. We show state-of-the-art segmentation performance on diverse datasets. The prediction is fast and takes less than a minute to run on a single process. In the future we intend to use more accurate imaging equations and simulate other pulse sequences to increase the range of application.
\vspace{-0.25em}
\bibliographystyle{splncs03}
\bibliography{amod-short-2016-03-14}

\begin{thebibliography}{10}
\providecommand{\url}[1]{\texttt{#1}}
\providecommand{\urlprefix}{URL }

\bibitem{asman2012tmi}
Asman, A.J., Landman, B.A.: Formulating spatially varying performance in the
  statistical fusion framework. IEEE TMI  31(6),  1326--1336 (June 2012)

\bibitem{buckner2004protocol}
Buckner, R., Head, D., Parker, et~al.: A unified approach for morphometric and
  functional data analysis in young, old, and demented adults using automated
  atlas-based head size normalization: reliability and validation against
  manual measurement of total intracranial volume. NeuroImage  23(2),  724 --
  738 (2004)

\bibitem{fischl2004aseg}
Fischl, B., Salat, D.H., et~al.: Whole brain segmentation: Automated labeling
  of neuroanatomical structures in the human brain. Neuron  33(3),  341 -- 355
  (2002)

\bibitem{Fischl2004}
Fischl, B., Salat, D.H., van~der Kouwe, A.J., et~al.: Sequence-independent
  segmentation of magnetic resonance images. NeuroImage  23(S1),  S69--S84
  (2004)

\bibitem{han2006ni}
Han, X., Jovicich, J., Salat, D., et~al.: Reliability of {MRI}-derived
  measurements of human cerebral cortical thickness: The effects of field
  strength, scanner upgrade and manufacturer. NeuroImage  32(1),  180 -- 194
  (2006)

\bibitem{jog2015media}
Jog, A., Carass, A., Roy, S., Pham, D.L., Prince, J.L.: {MR} image synthesis by
  contrast learning on neighborhood ensembles. MedIA  24(1),  63--76 (2015)

\bibitem{jovicich2013ni}
Jovicich, J., et~al.: Brain morphometry reproducibility in multi-center {3T
  MRI} studies: A comparison of cross-sectional and longitudinal segmentations.
  NeuroImage  83,  472 -- 484 (2013)

\bibitem{niftynet17}
Li, W., Wang, G., Fidon, L., Ourselin, S., Cardoso, M.J., Vercauteren, T.: On
  the compactness, efficiency, and representation of {3D} convolutional
  networks: Brain parcellation as a pretext task. In: IPMI (2017)

\bibitem{mugler1990mrm}
Mugler, J.P., Brookeman, J.R.: {Three-dimensional magnetization-prepared rapid
  gradient-echo imaging (3D MPRAGE)}. Mag. Reson. Med.  15(1),  152--157 (1990)

\bibitem{puonti2016samseg}
Puonti, O., Iglesias, J.E., Leemput, K.V.: Fast and sequence-adaptive
  whole-brain segmentation using parametric {B}ayesian modeling. NeuroImage
  143,  235 -- 249 (2016)

\bibitem{unet2015}
Ronneberger, O., Fischer, P., Brox, T.: U-net: Convolutional networks for
  biomedical image segmentation. In: MICCAI 2015. pp. 234--241 (2015)

\bibitem{shiee2010ni}
Shiee, N., {et al.}: {A topology-preserving approach to the segmentation of
  brain images with Multiple Sclerosis lesions}. NeuroImage  49(2),  1524--1535
  (2010)

\bibitem{wachinger2017}
Wachinger, C., Reuter, M., Klein, T.: Deep{NAT}: Deep convolutional neural
  network for segmenting neuroanatomy. NeuroImage  (2017)

\bibitem{wang2013malf}
Wang, H., Suh, J.W., Das, S.R., Pluta, J.B., Craige, C., Yushkevich, P.A.:
  Multi-atlas segmentation with joint label fusion. IEEE Transactions on
  Pattern Analysis and Machine Intelligence  35(3),  611--623 (March 2013)

\bibitem{wang2014mprage}
Wang, J., He, L., Zheng, H., Lu, Z.L.: Optimizing the magnetization-prepared
  rapid gradient-echo ({MP-RAGE}) sequence. PloS one  9(5),  e96899 (2014)

\end{thebibliography}

\end{document}